\documentclass[letterpaper, 10 pt, conference]{ieeeconf}  

\IEEEoverridecommandlockouts                              

\usepackage{subcaption}
\usepackage{url}
\usepackage{graphicx} 
\usepackage{amsmath} 
\usepackage{cite}

\title{\LARGE \bf
Learning to Centralize Dual-Arm Assembly
}

\author{Marvin Alles and Elie Aljalbout$^{1}$
\thanks{$^{1}$Technical University of Munich, 80797 Munich, Germany
        {\tt\small \{firstname.lastname@tum.de\}}}%
}

\begin{document}

\maketitle
\thispagestyle{empty}
\pagestyle{empty}


\begin{abstract}
	Robotic manipulators are widely used in modern manufacturing processes. However, their deployment in unstructured environments remains an open problem. To deal with the variety, complexity, and uncertainty of real-world manipulation tasks, it is essential to develop a flexible framework with reduced assumptions on the environment characteristics. In recent years, reinforcement learning (RL) has shown great results for single-arm robotic manipulation. However, research focusing on dual-arm manipulation is still rare. From a classical control perspective, solving such tasks often involves complex modeling of interactions between two manipulators and the objects encountered in the tasks, as well as the two robots coupling at a control level. Instead, in this work, we explore the applicability of model-free RL to dual-arm assembly. As we aim to contribute towards an approach that is not limited to dual-arm assembly, but dual-arm manipulation in general, we keep modeling efforts at a minimum. Hence, to avoid modeling the interaction between the two robots and the used assembly tools, we present a modular approach with two decentralized single-arm controllers which are coupled using a single centralized learned policy. We reduce modeling effort to a minimum by using sparse rewards only. Our architecture enables successful assembly and simple transfer from simulation to the real world. We demonstrate the effectiveness of the framework on dual-arm peg-in-hole and analyze sample efficiency and success rates for different action spaces. Moreover, we compare results on different clearances and showcase disturbance recovery and robustness, when dealing with position uncertainties. Finally we zero-shot transfer policies trained in simulation to the real world and evaluate their performance. Videos of the experiments are available at the project website: \url{https://sites.google.com/view/dual-arm-assembly/home} 
	
	\end{abstract}


\section{INTRODUCTION}

\begin{figure}[!t]
	\centering
	\begin{subfigure}[t]{0.43\textwidth}
		\centering
		\includegraphics[width=0.9\linewidth]{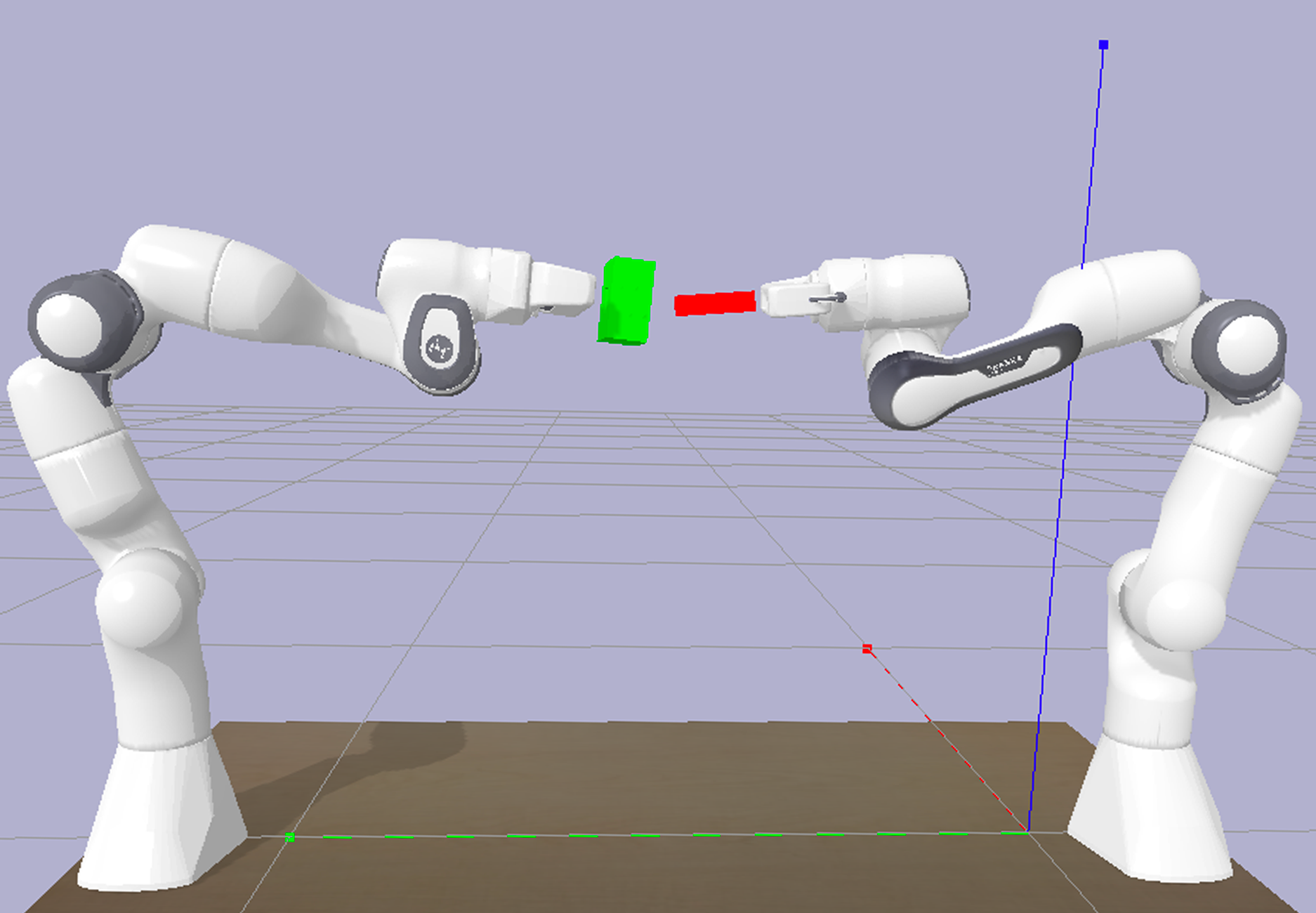}
		\subcaption{Simulation}
		\label{fig:simulation}
	\end{subfigure}
	
	\begin{subfigure}[t]{0.43\textwidth}
		\centering
		\includegraphics[width=0.9\linewidth]{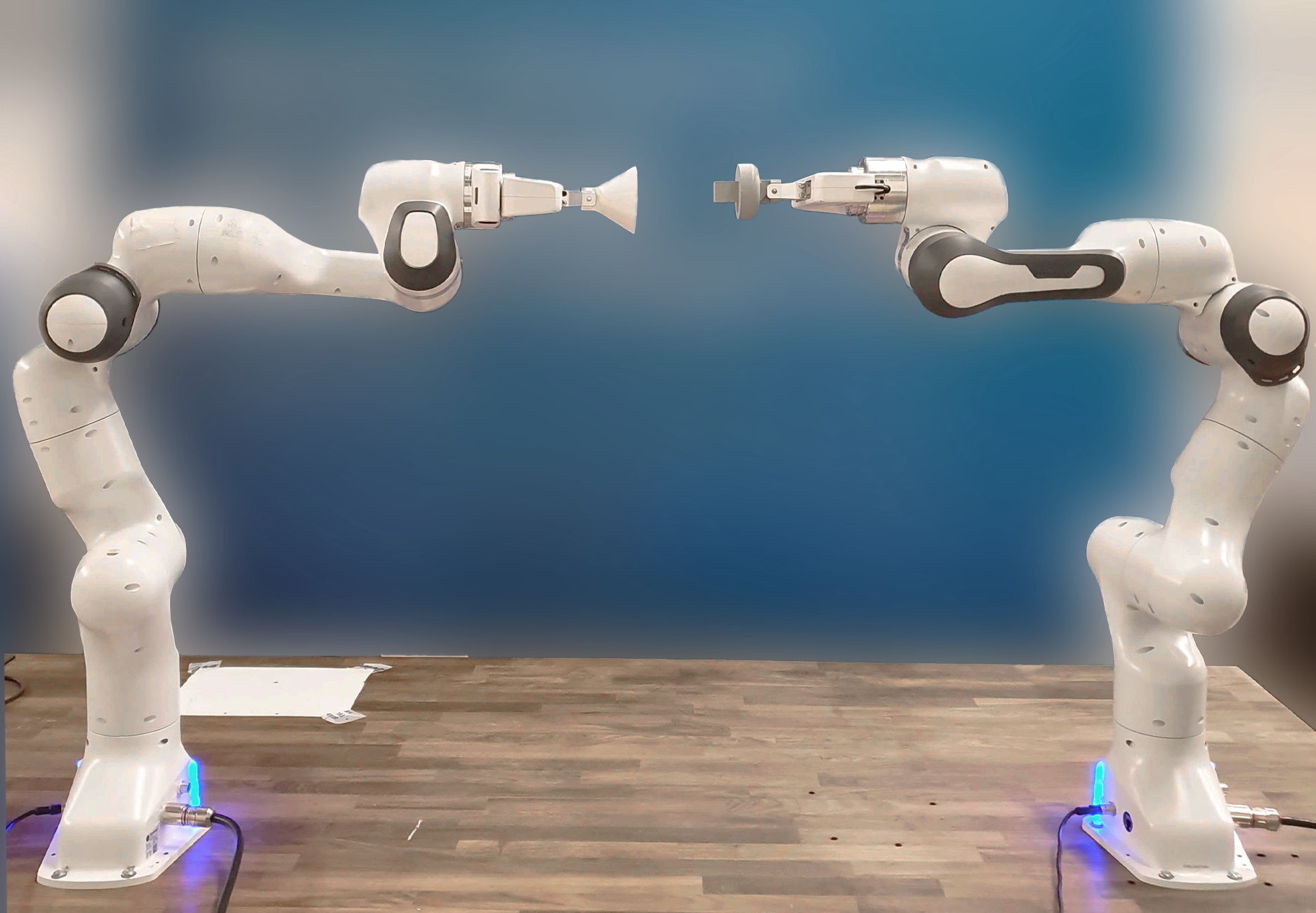}
		\subcaption{Real-world}
		\label{fig:real_world}
	\end{subfigure}
	
	\caption{Simulation-to-real transfer: The policy is trained in simulation (a) to perform dual-arm peg-in-hole, and transferred to the real-world (b) without additional training.}
\label{fig:sim2real}
\end{figure}

In recent years, robotic manipulation has been an active field of research. However, work focusing on dual-arm manipulation is still rare and limited. A second robotic arm enhances dexterity but also introduces new challenges and extra modeling efforts, such as additional degrees of freedom and interactions between manipulators. Thus, it is common to use a complex task-specific control structure with multiple control loops. However, such methods are usually restricted to certain classes of tasks and often assume access to accurate models of the objects involved in the task as well as the interaction dynamics between the two robots. In this work, we focus on task-agnostic methods for dual-arm manipulation. Hence, throughout this work, we restrict task-related modeling to a minimum. As real world-robot learning experiments could be very expensive, we only experiment with dual-arm assembly, but restrict ourselves from including or modeling any kind of knowledge specific to this task. 

To this end, deep reinforcement learning (RL) is a promising approach to tackle this problem. Thereby manipulation tasks can be learned from scratch by interaction with the environment. However, deep RL alone would require a lot of training samples which are expensive to collect in a real-world setup. Instead, it is preferable to introduce inductive biases into our architecture as to facilitate the learning process. Namely, we train a policy network only to generate high-level trajectories and use well-established control techniques to track those trajectories. Such a modular architecture also allows for zero-shot sim-to-real transfer. This enables us to do all the training in simulation. 

In general, we distinguish  between decentralization and centralization on both the control level and policy level. On a control level, centralized control requires large modeling efforts and is not task-agnostic, our work considers a decentralized approach. With that in mind, two general paradigms are conceivable: the first one involves two separate decoupled RL agents which can be trained in a multi-agent RL setting and the second one utilizes a single policy controlling both arms. The latter is more feasible as it couples control of both manipulators through a policy network, resulting in an overall centralized method, and, thus increases precision and efficiency. Our method is based on the latter approach and attempts to learn a single policy using off-policy RL. Intuitively, such an approach can be thought of as a way to centralize decentralized control based on RL. 

This paper aims at exploring the applicability of deep RL to dual-arm assembly. Hence, we propose a framework to learn a policy for this task based on a combination of recent advances in RL and well-established control techniques. To reduce the need for task-specific knowledge and to avoid introducing additional bias in the learning process, we test our methods solely with sparse rewards. Nevertheless, only receiving a reward after successfully solving the task provides less guidance to the agent as no intermediate feedback is given. Thus, the already challenging task of dual-arm manipulation becomes more complicated and sample-inefficient. That is why we rely on simulation to train our policy and transfer the results to the real world (figure~\ref{fig:sim2real}). Moreover, we design our framework with the goal of avoiding elaborate sim-to-real transfer procedures.

To demonstrate the effectiveness of our approach we evaluate our method on a dual-arm peg-in-hole task, as it requires high dexterity to manipulate two objects with small clearances under consideration of contact forces. We first use PyBullet~\cite{coumans2019} to create a real-time physics simulation environment and analyze the proposed approach with a focus on sample efficiency, performance and robustness. We then test the learned behavior in a real-world setup with two Franka Emika Panda robots and demonstrate the feasibility of our method under minimized sim-to-real transfer efforts. Our contributions can be summarized as follows:
\begin{itemize}
	\item We explore and formulate a new paradigm for learning dual-arm assembly tasks.
	\item We compare the performance of different action spaces and controllers on the success and robustness of the learned policies.
	\item We show that it's possible to zero-shot transfer policies trained in simulation to the real-world, when using a suitable and abstract action space.
	\item To our knowledge, our work is the first to explore the applicability of model-free RL to contact-rich dual-arm manipulation tasks.
\end{itemize}

\section{RELATED WORK}

\begin{figure*}[!thpb]
	\includegraphics[width=1\linewidth]{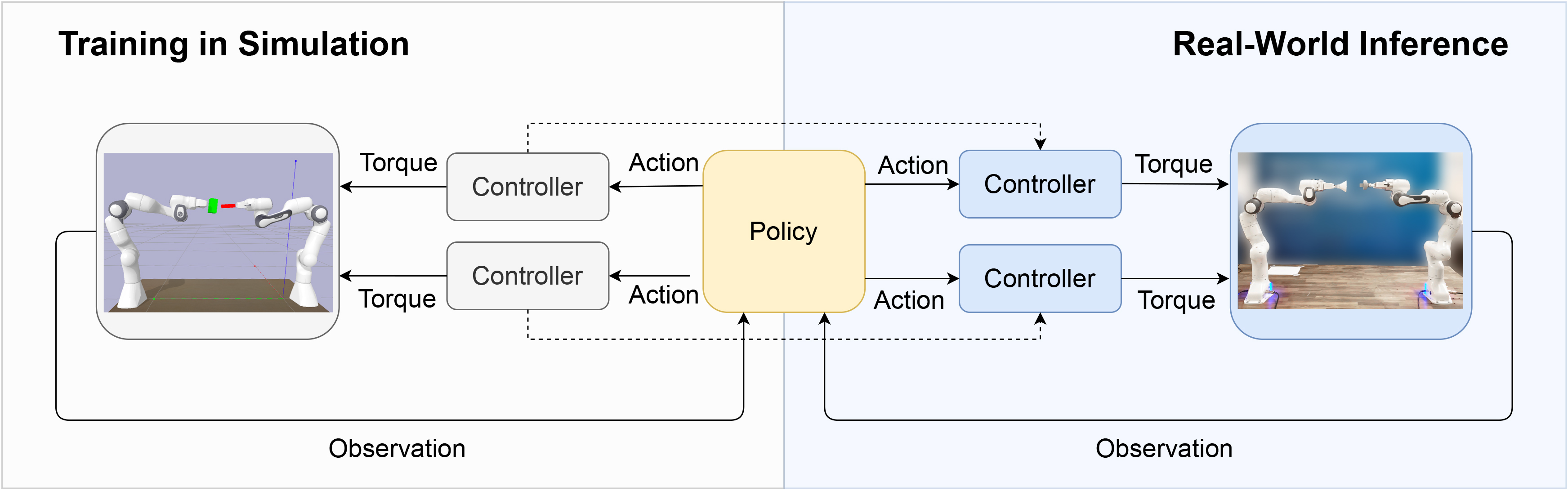}
	\caption{The diagram gives an overview of our method by showing the interaction between policy, controllers, robots and the environments (simulation and real-world). Our framework uses two decentralized single-arm controllers to avoid modeling the task-specific system dynamics and relies on a centralized policy to account for the overall interaction. Based on this architecture, we train our policy in simulation and zero-shot transfer it to the real-world environment. The sim-to-real transfer requires only minor adaptations of the controller parameters, as illustrated by the dashed line.}
	\label{fig:overview}
\end{figure*}

Dual-arm manipulation is a challenging area of research, which can be divided into decentralized and centralized approaches. The first one utilizes independent controllers for each robot with explicit~\cite{Petitti2016} or implicit~\cite{Wang2016,Tsiamis2015} communication channels and is often combined with leader/follower behavior~\cite{suomalainen2019improving, Wang2016}. Despite the resulting improvements in scalability and variability, decentralized control hardly reaches the efficiency and precision of centralized control, which integrates the control of both manipulators in a central unit. Among feasible manipulation objectives, peg-in-hole insertion can be seen as a benchmark, since it requires accurate positioning, grasping, and handling of objects in contact-rich situations. Therefore, we select the task of dual-arm peg-in-hole to evaluate the performance of our approach.

\textbf{Single-arm peg-in-hole.} As research focusing on dual-arm peg-in-hole assembly is rare and mostly limited to extensive modeling~\cite{suomalainen2019improving, Park2014,Zhang2017}, research on classical peg-in-hole assembly with a single robotic arm provides a perspective on model-free approaches based on reinforcement learning. \cite{Vecerik2017} and~\cite{Schoettler2019} show that sparse rewards are sufficient to successfully learn a policy for an insertion task if combined with learning from demonstrations. The work in~\cite{Schoettler2019} uses residual reinforcement learning to leverage classical control, which performs well given sparse rewards only and provides a hint, that the choice of action space can be crucial. An evaluation of action spaces on the task of single-arm peg-in-hole with a clearance of $2mm$ and a shaped reward function is presented in~\cite{Varin2019}, where cartesian impedance control performs best. Moreover, \cite{Beltran-Hernandez2020} applies position-force control with model-free reinforcement learning for peg-in-hole with a focus on transfer-learning and domain randomization.

\textbf{Decentralized Dual-Arm Manipulation.} In the work by~\cite{suomalainen2019improving}, a decentralized approach for the dual-arm peg-in-hole task is proposed. The method is based on a leader/follower architecture. Hence, no explicit coupling between both manipulators is required. The leader would perform the insertion, and the follower would hold its position and be compliant with the applied forces. Similar to the previously mentioned work, \cite{Zhang2017} utilizes a decentralized approach, where the hole keeps the desired position with low compliance and the peg is steered in a spiral-screw motion towards insertion with high compliance.
But despite reducing the necessity to model their interaction, both approaches lack dexterity, i.e., there is only one robot actively acting in the environment. In a general pipeline, there should be enough flexibility for both arms to be actively contributing towards the objective. Furthermore, \cite{Park2014} presents a method based on decomposing the task into phases and utilizing a sophisticated control flow for the whole assembly process. Despite reducing efforts in modeling the interaction explicitly, the control flow is only engineered for one specific task and lacks dexterity as movements are bound to the preprogrammed procedure.

\textbf{Centralized Dual-Arm Manipulation.}
Work on centralized dual-arm manipulation focuses on cooperative single object manipulation, hence the applicability is limited to a few use-cases.
\cite{pairet2019learning} propose a learning-based approach and evaluate their method on a synthetic pick-and-place task. A set of primitive behaviors are demonstrated to the robot by a human, the robot combines those behaviors and tries to solve a given task. Finally, an evaluator measures its performance and decides if further demonstrations are necessary. The approach has promising potential towards more robust and less task-specific dual-arm manipulation. However, besides the required modeling efforts, it is limited by the human teaching process, which introduces an additional set of assumptions, limiting its applicability to semi-structured environments. 
Besides that, classical methods for cooperative single object manipulation with centralized control highly rely on accurate and complex modeling of the underlying system dynamics.~\cite{Caccavale2001,Caccavale2008,Erhart2013,Heck2013,Ren2015,Bjerkeng2014}

\textbf{Sim-to-Real Transfer.} Sample inefficiency is one of the main challenges of deep RL algorithms. The problem is even worse for robotic tasks, which involve high-dimensional states and actions as well as complex dynamics. This motivates the use of simulation for data collection and training. However, due to the inaccuracies in the physics modeling and image rendering in simulation, policies trained in simulation tend to fail in the real world. This is usually referred to as the "reality gap". The most popular paradigm to approach this problem is domain randomization~\cite{8202133}. The main goal of domain randomization is to subject the agent to samples based on diverse simulation parameters concerning the object~\cite{8202133} and the dynamics properties~\cite{peng2018sim}. By doing so, the learned policy is supposed to be able to generalize to the different physical properties of real-world tasks. Recent work has explored active parameters sampling strategies as to dedicate more training time for troublesome parameter settings~\cite{mehta2020active}. Another approach for sim-to-real transfer is system modularity~\cite{clavera2017policy}. Here a policy is split into different modules responsible for different objectives such as pose detection, online motion planning and control. Only components that won't suffer from the reality gap are trained in simulation and the rest is adapted or tailor-made for the real-world setup. This comes in contrast to the most common end-to-end training in deep RL~\cite{levine2016end}. In our work, we use a modular architecture to enable zero-shot sim-to-real transfer. Namely, we parameterize the controllers differently in simulation compared to the real world to allow using the same high-level policy network.

Despite various contributions towards a general framework for dual-arm manipulation, we do not know of any work that successfully fulfills all requirements. Therefore, this work aims at proposing a unified pipeline for dual-arm manipulation based on a minimal set of assumptions. To the best of our knowledge, no prior work exists on contact-rich dual-arm peg-in-hole with model-free reinforcement learning, nor centralized dual-arm control for non-single object manipulation tasks in general.

\section{LEARNING TO CENTRALIZE}

In this section, we introduce our framework for dual-arm manipulation. We intend to reduce the required modeling effort to a minimum, which is why our approach is based on model-free reinforcement learning with sparse rewards. The approach does not require a specific dual-arm controller, since control is only coupled at a policy level.
\subsection{Controller}
The classical approach in centralized dual-arm control strategies is to model the manipulator's and the object's dynamics explicitly and achieve coupling of both robotic arms by using multiple control loops. The system dynamics of the $i$-th manipulator can be described by joint space equations of motion:
\begin{equation}
M_i({q_i}) \ddot{{q_i}}+{C_i}({q_i}, \dot{{q_i}}) \dot{{q_i}}+{d_i}({q_i}, \dot{{q_i}})+{g_i}({q_i})={\tau_i}-\tau_{ext_i}
\label{eq:sys_dyn}
\end{equation}
\begin{equation}
\tau_{ext_i}={J_i}^{T}({q_i}) {h_i}
\label{eq:external_wrench}
\end{equation}
Where $q_i$ is the vector of joint positions, $M_i(q_i)$ is the inertia matrix, $C_i(q_i, \dot q_i)$ is the Coriolis/centrifugal matrix, $g_i(q_i)$ is gravity vector, $\tau_i$ is the vector of joint torques and $\tau_{ext_i}$ represents external torques, which can be further decomposed into the external wrench $h_i$ as in~(\ref{eq:external_wrench}). As both robots interact with each other directly or through manipulation of objects, the respective external wrenches have to include all forces and moments which are applied to the robot. Hence, to create a valid dynamic chain, the rigid-body dynamics of the j-th object are described by the following equation: 
\begin{equation}
M_j \ddot{x_j} + C_j(\dot{x_j})\dot{x_j} + g_j = h_j - h_{ext_j}
\label{eq:obj_dyn}
\end{equation}
where $M_j$ is the inertia matrix, $C_j$ is the coriolis/centrifugal matrix, $g_j$ is the gravity vector, $h_j$ the wrench exerted on the environment and $h_{ext_j}$ the  external wrench exerted by a e.g. manipulator.
In case that manipulators and objects form a closed dynamical chain, the system dynamics can be described by concatenating the respective external wrenches. Thereby, the dynamic equations need to be adapted to the specific manipulation task, but are not universally valid. For example single object manipulation with two manipulators can be described as follows:
\begin{equation}
h_{object} \approx G_1 h_{robot1} + G_2 h_{robot2}
\label{eq:single_obj}
\end{equation} 
Whereas, dual-arm peg-in-hole could be defined by:
\begin{equation}
\begin{aligned}
h_{peg} \approx G_1 h_{robot1} \\
h_{hole} \approx G_2 h_{robot2} \\
ContactDynamicsModel(h_{peg}, h_{hole}) \\
\end{aligned}
\label{eq:peg_in_hole}
\end{equation}
With $G_i$ as respective grasp matrices, $ContactDynamicsModel$ as placeholder for contact modeling and of course with strongly simplified assumptions (e.g. neglecting mechanical stresses, no environment interactions, assuming force closure, neglecting object geometries, etc).

Under considerations of simplified assumptions and various constraints, centralized single object manipulation can be tackled by classical control: Based on the dynamics model, commonly a hierarchical control strategy through multiple loops is applied, where the outer loops realize the main objective such as desired object movements and the inner loop accounts to generate a firm grasp and bounded internal forces~\cite{Caccavale2001,Caccavale2008,Erhart2013,Heck2013,Ren2015,Bjerkeng2014}.
The particular control loops can utilize any control strategy. Nevertheless, impedance control is a common choice to achieve compliant behavior~\cite{Caccavale2001,Caccavale2008,Heck2013,Ren2015}, as contact forces are limited by coupling the control torque with position $p$ and velocity $v$~(\ref{eq:impedance_control_general}). The control torque is calculated by multiplying the gains $K_p$ and $K_v$ with the difference of desired and actual position and velocity respectively:
\begin{equation}
\tau = f_0(K_p (p_{des} - p) + K_v (v_{des} - v)) + f_1()
\label{eq:impedance_control_general}
\end{equation}
The principle can be applied in joint space or task space. $f_0$ and $f_1$ are generic functions to account for the variety of control laws and their additions (e.g. equation~\ref{eq:cartesian_imp_control}). Besides the success in dual-arm cooperative single object manipulation tasks, to our knowledge, no method exists for solving general assembly tasks with centralized control. Methods to explicitly model contacts and complex geometries are so far not powerful enough to formulate closed-loop dynamics and use a centralized controller. Even if control by classical methods would be feasible, an explicit set of dynamic equations, constraints and adapted control loops for each task would be required. Hence, a general framework for dual-arm robotic manipulation needs to be based on a different approach.

An alternative way to bridge the gap is to use learning-based methods. Especially learning through interactions with the environment as in deep RL provides a promising way to facilitate complex manipulation tasks without the need for labeled data or human demonstrations. A learning signal is solely provided by rewarding the intended behavior. Deep RL then tries to maximize the accumulated reward, leading to policies/controllers with are compliant with the incentivized behavior.

Our approach is based on the idea of combining classical control with deep RL: We merge a policy network as high-level control and two independent low-level controllers. We thereby dispose of the need of designing a coupled control method in the classical sense. The controllers can be designed in a straightforward way without the need for purpose-built dual-arm control algorithms, allowing the use of any single-arm action space. The policy learns to inherently compensate for the constraints resulting from dual-arm interactions and provides individual action inputs for each controller. Furthermore, coupling at policy level is convenient to implement as the policies action space can simply be extended to include a second controller. The overall system is illustrated in figure~\ref{fig:overview}.

Besides the so far mentioned advantages of our framework, the method enables improved sim-to-real transfer: Both low-level controllers can be adjusted to the specifics of the real world, whereas the high-level policy can be zero-shot transferred from simulation. Thereby, the need for expensive and difficult real-world experiments can be reduced to a minimum and the benefits from classical methods and learning-based approaches are combined.

\subsection{Action Space}
\label{sec:actispace}
Classical control approaches for manipulation are often based on impedance control, as it comes with the previously mentioned advantages. However, since our method tries to compensate for interaction forces at a trajectory level, we explore different control laws as action spaces and study their effect on the task success. \vspace{2pt}
\\
\textbf{Joint position control.} 
First of all, we use joint position control (equation~\ref{eq:joint_position_control}) to compute a torque command: Both gains, $k_p$ and $k_v$, are set to a constant value, $q_{actual}$ and $\dot{q}_{actual}$ evaluated at run-time and $q_{desired}$ inferred by the policy.
\begin{equation}
\tau = k_p \cdot (q_{desired} - q_{actual}) + k_v \cdot \dot{q}_{actual}
\label{eq:joint_position_control}
\end{equation}
\textbf{Cartesian impedance control.}
Second, we implement cartesian impedance control~\cite{impedancecontrol}: The action space allows to move control from joint space to cartesian space and includes model information such as the cartesian inertia matrix $\Lambda(x)$ and the jacobian matrix $J(q)$ as well as the gravity compensation term $\tau_{gc}$. As the degrees of freedom exceed the number of joints, nullspace compensation $\tau_{ns}$ is added. Instead of $x_{desired}$, $\Delta x=x_{desired} - x_{actual}$ is directly passed in as action input.
\begin{equation}
\begin{split}
\tau = J(q)^T\Lambda(x) \cdot (k_p \cdot \Delta x 
+ k_v \cdot \dot{x}_{actual}) + \tau_{gc} + \tau_{ns}
\label{eq:cartesian_imp_control}
\end{split}
\end{equation}
\textbf{Variable cartesian impedance control.}
Variable cartesian impedance control~\cite{Martin-martin} is based on classical cartesian impedance control, though adds $k_p$ to the action space making control more variable to react with higher or lower stiffness if needed. We use anisotropic gains and couple velocity gains via $k_v = 2 \sqrt{k_p}$ to achieve critical damping.

\subsection{Reinforcement Learning}
In our method, the policy is responsible for generating high-level trajectories, which are later-on tracked by the chosen controller.
We train the policy network using model-free RL. The policy receives the robot states to infer a control signal (action) for the above mentioned control laws (action spaces). We combine joint positions $q_i$, joint velocities $\dot{q}_i$ and joint torques $\tau_i$ of both robotic arms respectively, as well as cartesian positions and orientations of the end-effectors as state input $s=[q_0,\dot{q}_0,\tau_0,ee_{pos_0},ee_{ori_0},q_1,\dot{q}_1,\tau_1,ee_{pos_1},ee_{ori_1}]$. Nevertheless, the state might need to be adjusted if the framework is applied for a different task, which for instance includes additional objects. 

The proposed method is not restricted to a specific model-free RL algorithm, though an off-policy algorithm is desirable to facilitate high sample efficiency and allow the use of experience replay. Thus, we use Soft Actor-Critic (SAC)~\cite{Haarnoja2018_2} as the algorithm presents state of the art performance and sample efficiency, but could potentially be replaced by others such as Deep Deterministic Policy Gradients (DDPG)~\cite{Lillicrap2016} or Twin Delayed Deep Deterministic Policy Gradients (TD3)~\cite{Fujimoto2018}.

To enhance sample efficiency, the environment is implemented in a goal-based way. Thereby, the achieved goal $goal^{achieved}$ is returned alongside the environment state and can easily be compared to the desired goal $goal^{desired}$ to compute a reward $r$. Reward engineering is not necessary as we use a sparse reward~(\ref{eq:reward}).
\begin{equation}
r = \left\{\begin{array}{ll}1, & \text { if } \sum_{i=1}^{n}\left|goal^{achieved}_i - goal^{desired}_i\right| < \delta \\ 0, & \text { else }\end{array}\right.
\label{eq:reward}
\end{equation}
In~\cite{Zuo2020} a similar setup is combined with the concept of Hindsight Experience Replay (HER)~\cite{Andrychowicz2017} for the single-arm robotic manipulation tasks push as well as pick-and-place. Their results point out, that HER is sufficient to enhance the performance if only sparse rewards are available.
Hence, to counteract the more challenging training task when using sparse compared to dense rewards, we use HER to augment past experiences. By replaying experiences with goals that have been or will be achieved along a trajectory the agent shall generalize a goal-reaching behavior. Hence, unsuccessful experiences still help to guide an agent, as a sparse reward otherwise does not provide feedback on the closeness to the desired goal.

\begin{figure*}[!thpb]
	\centering
	\begin{subfigure}[t]{0.6\textwidth}
		\centering
		\includegraphics[width=1\linewidth]{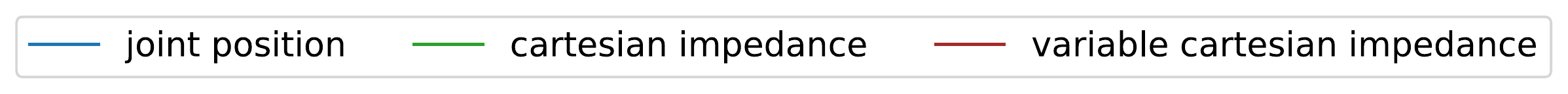}
	\end{subfigure}
	\begin{subfigure}[t]{0.49\textwidth}
		\centering
		\includegraphics[width=1\linewidth]{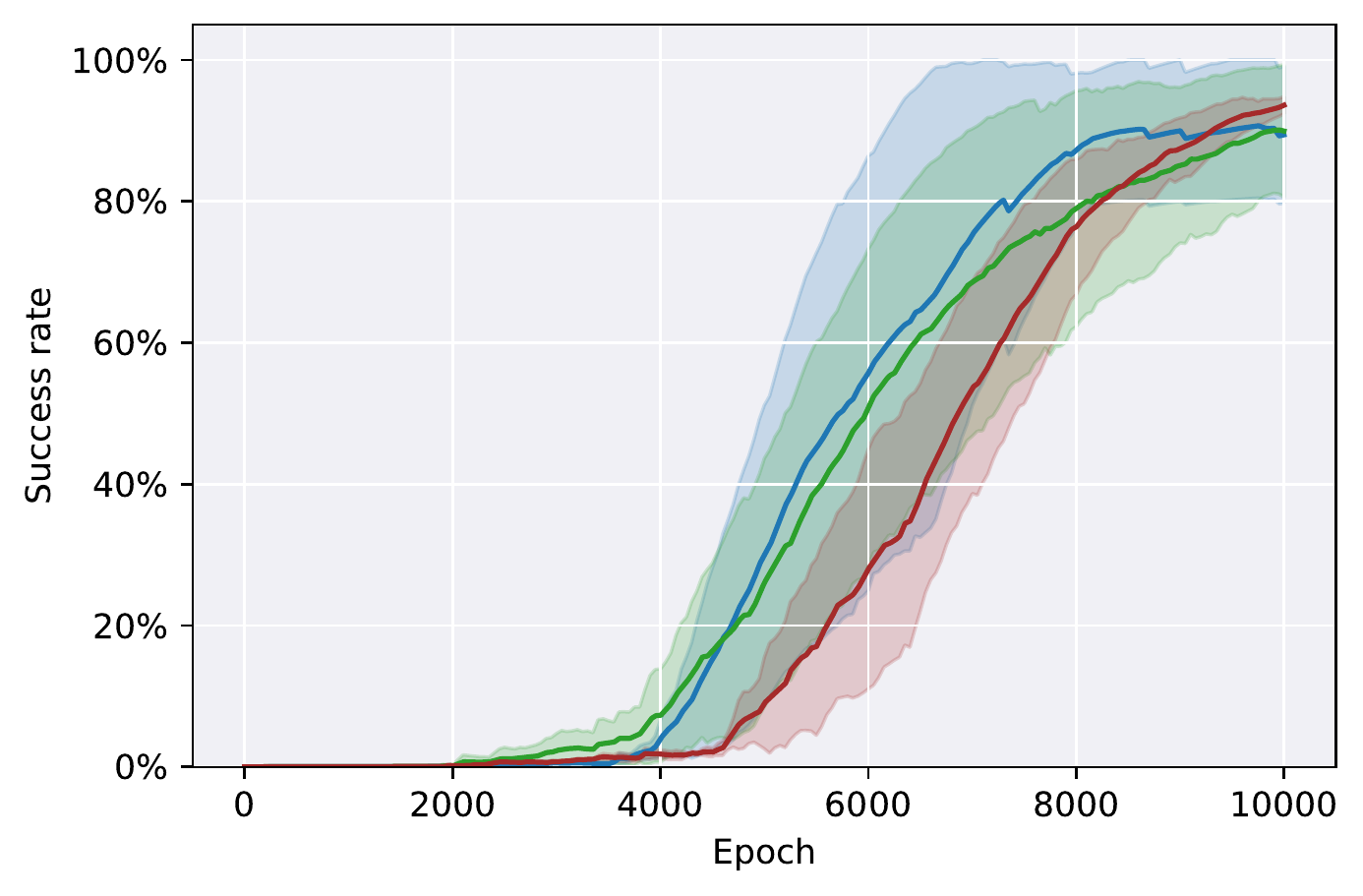}
		\subcaption{$2mm$ clearance}
		\label{fig:sim_2mm}
	\end{subfigure}
	\begin{subfigure}[t]{0.49\textwidth}
		\centering
		\includegraphics[width=1\linewidth]{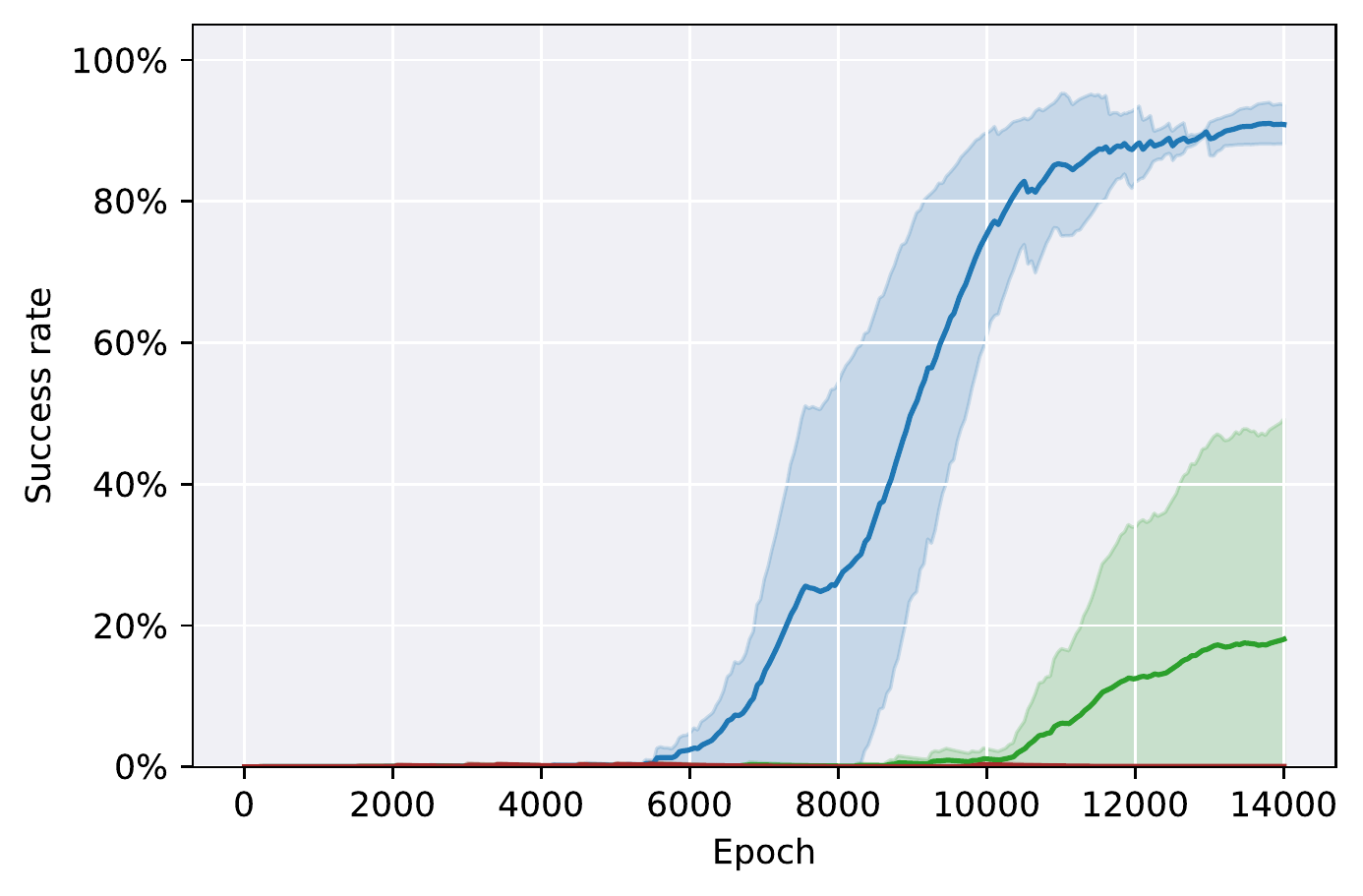}
		\subcaption{$0.5mm$ clearance}
		\label{fig:sim_05mm}
	\end{subfigure}
	
	\caption{Training results in simulation using different action spaces (joint position control, cartesian impedance control and variable cartesian impedance control) on the task of dual-arm peg-in-hole assembly separated for different clearances (2mm (a) and 0.5mm (b)).}
	\label{fig:results_action_space}
\end{figure*}

\subsection{Implementation}
We implement the general training and evaluation procedure in the following way: During each epoch, one full episode is gathered by interacting with the environment followed by $1000$ optimization steps. Moreover, we calculate the success rate every fifth epoch by averaging over $10$ test cycles. We use ADAM~\cite{kingma2017adam} as optimizer with a learning rate of $1\mathrm{e}{-5}$ and a batch size of $256$. The experience replay memory size is set to $800k$ and training starts after storing $10000$ samples. The q-network and policy-network consist of $4$ and $3$ linear layers respectively with a hidden dimension of $256$ and ReLU~\cite{Agarap2018DeepLU} activation functions. To update the target networks, we set an updating factor of $0.005$. HER is set to use the strategy "future" with sampling of $6$ additional experiences~\cite{Andrychowicz2017}.
All hyper-parameters are tuned manually and kept fixed for all experiments.

\section{Experimental Setup}
We design experiments to answer the following questions: 
\begin{itemize}
	\item Can a central policy successfully learn dual-arm manipulations skills based on a decentralized control architecture?
	\item What action space leads to the highest success rate and the most robust policies?
	\item Is our method robust against disturbances and position uncertainty?
	\item Is a modular design enough to zero-shot transfer policies from simulation to the real-world?
\end{itemize}  

To answer these questions, we evaluate the proposed method on the task of peg-in-hole assembly with two Franka Emika panda manipulators both in simulation (figure~\ref{fig:sim2real}A) and in the real world (figure~\ref{fig:sim2real}B). The simulation environment is created using PyBullet~\cite{coumans2019}. We design it to match the real-world setup as closely as possible. Both setups are similar except for the environment frequency which is 240Hz in simulation and 1KHz in the real world. The policy is operating at 60Hz. Nevertheless, the robots only obtain the last state for control. To exclude the process of gripping and restrict movements of peg and hole, both are fixed to the end-effector of the respective robot arm. In the real-world experiments, this corresponds to the peg and hole being attached to the gripper of each manipulator. Furthermore, to enable an evaluation with increasing difficulty, pairs of pegs and holes have been created with a clearance of $2mm$ and $0.5mm$. Moreover, we define the goal state as the relative distance between the positions of peg and hole. 
Both robots start with a significant distance and varying orientation with an offset around the initial joint position of $q_{init} = [0.0, -0.54, 0.0, -2, -0.3, 3.14, 1.0]$. We restrict robot movements by their joint limits, whereas the workspace is not bounded. Furthermore, the robot bases are positioned on a table with a distance of $1.3m$ and both oriented to the same side.
We use PyTorch~\cite{pytorch} to implement the models and train them on a single workstation equipped with a NVIDIA GeForce RTX 2080 GPU. 

\section{SIMULATION RESULTS}
We use the simulation to train the policy as well as to perform ablation studies and robustness tests. As can be seen in the supplementary video\footnote{\url{https://sites.google.com/view/dual-arm-assembly/home}}, the policy can be trained in simulation to learn a successful peg-in-hole insertion strategy. Both manipulators move equally towards each other without any bigger diversion. Furthermore, both are evenly involved in the process of aligning the end-effectors and pushing the peg inside. Hence, the approach does not lead to any leader/follower behavior where one end-effector just keeps its position similar to a single-arm solution.

We design the experiments to start with an offset of $\pm10\%$ from the initial joint positions and average all results over 4 seeds. All successful runs up to a clearance of $0.5mm$ converge to a success rate above $90\%$ in-between $10000$ and $14000$ epochs (figure~\ref{fig:results_action_space}). As we use sparse rewards, no intermediate information about the task success is available. Hence, the success rates of a single run tend to converge either to $100\%$ or stay at $0\%$, which can be easily seen for cartesian impedance control and a clearance of $0.5mm$, where only 1 out of 4 runs converges in the given time frame. Overall, variance and training duration is increasing with smaller clearances, which confirms that the task becomes more challenging as clearances decrease. 

\begin{figure*}[!thpb]
	\centering
	\begin{subfigure}[t]{0.8\textwidth}
		\centering
		\includegraphics[width=1\linewidth]{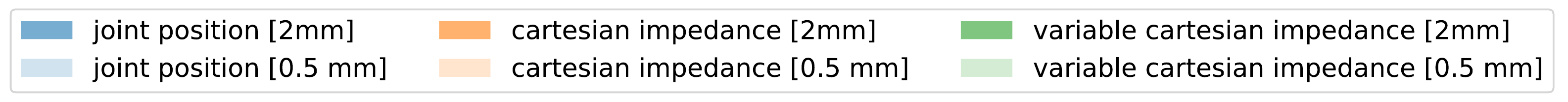}
	\end{subfigure}
	\begin{subfigure}[t]{0.49\textwidth}
		\centering
		\includegraphics[width=1\linewidth]{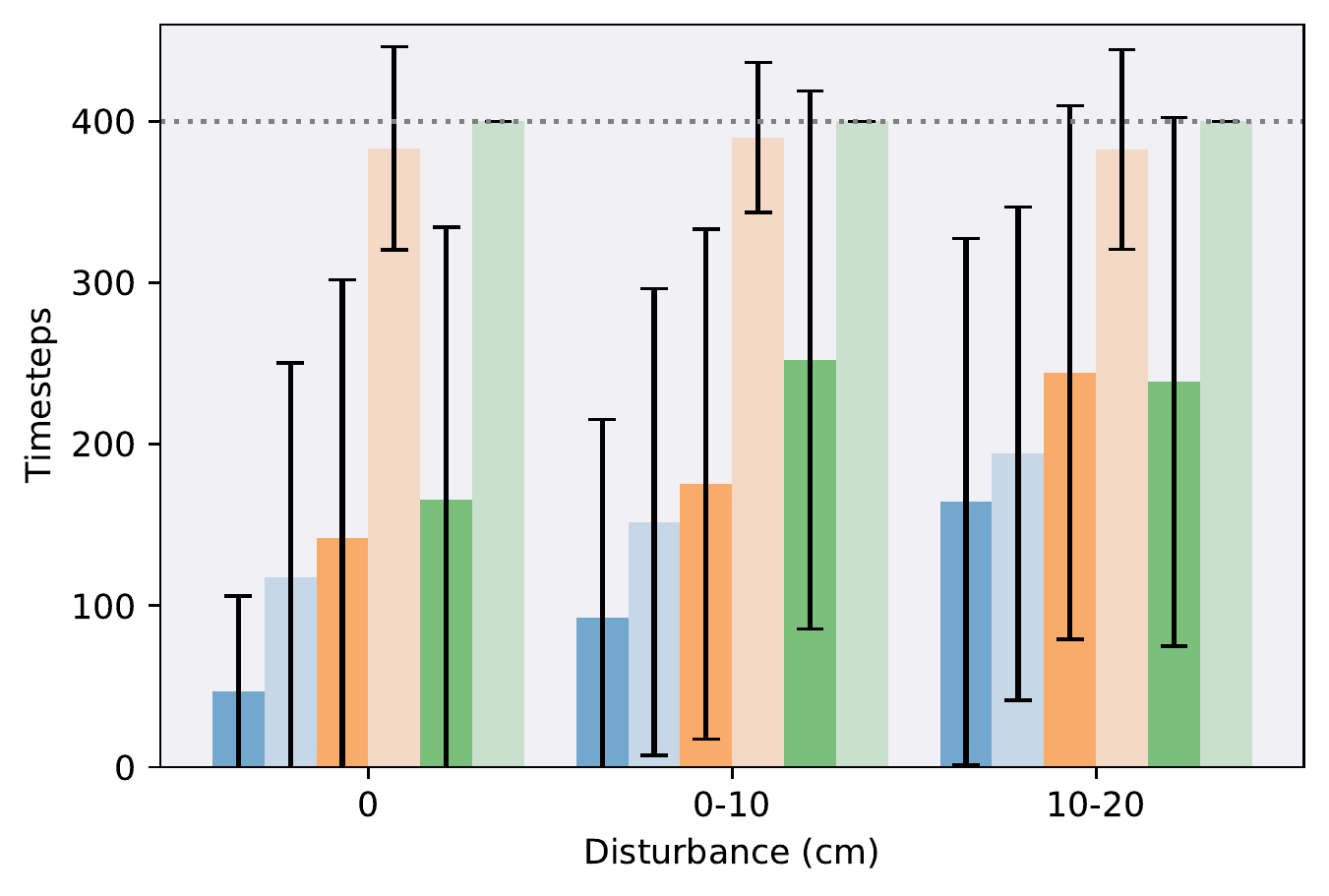}
		\subcaption{}
		\label{fig:external_force}
	\end{subfigure}
	\begin{subfigure}[t]{0.49\textwidth}
		\centering
		\includegraphics[width=1\linewidth]{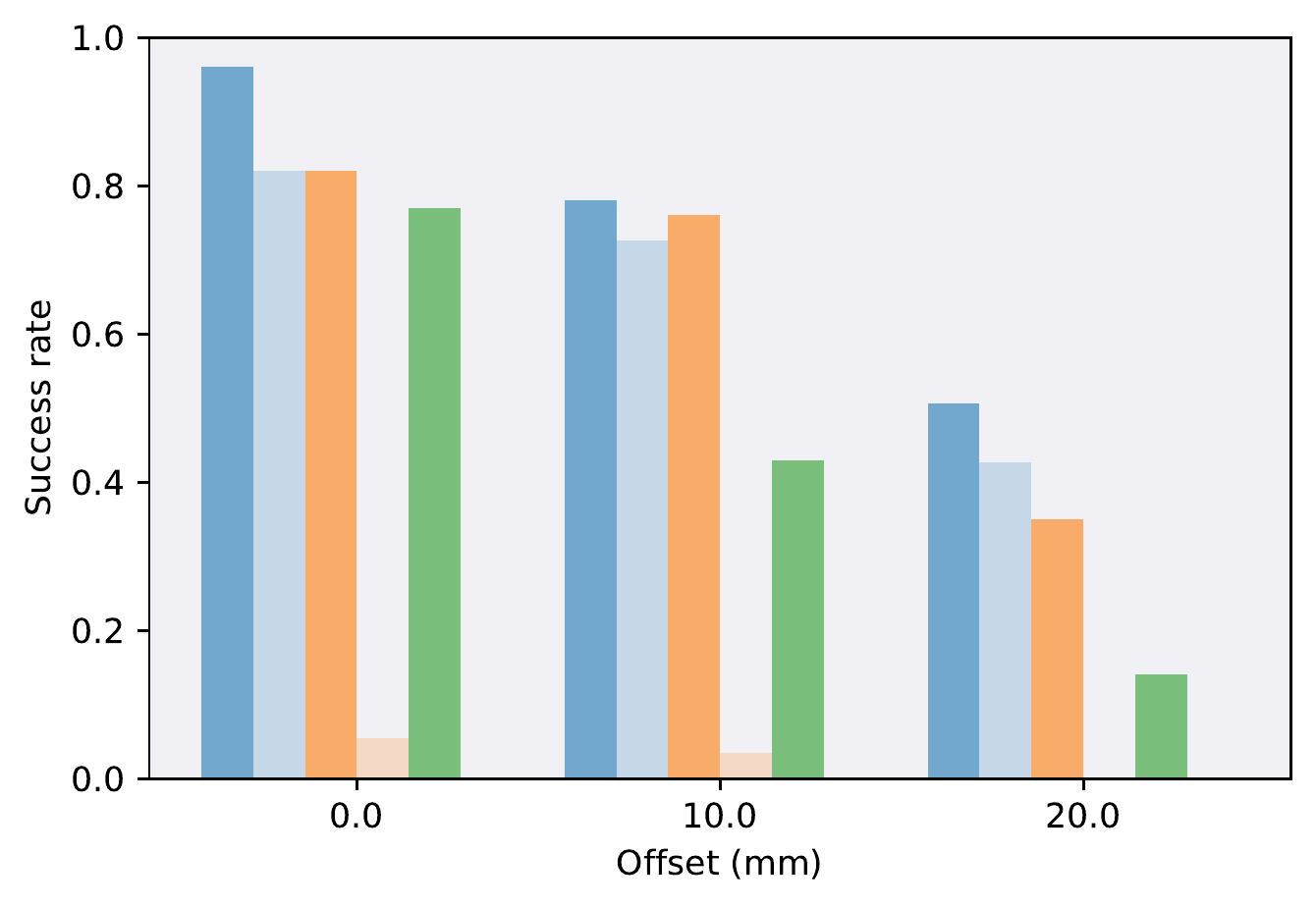}
		\subcaption{}
		\label{fig:random_pos}
	\end{subfigure}
	
	\caption{Robustness when using different action spaces (joint position control, cartesian impedance control and variable cartesian impedance control) on the task of dual-arm peg-in-hole assembly in simulation separated for different clearances. (a) Average episode duration's when applying a random disturbance to the peg during a fixed time frame in the beginning of each episode. (b) Average success rates when applying a random offset to the peg.}
	\label{fig:results_robustness}
\end{figure*}

\subsection{Ablation Studies}
We compare the following control variants to study the effect of different action spaces: joint position control, cartesian impedance control, and variable cartesian impedance control as introduced in section~\ref{sec:actispace}. Figure~\ref{fig:results_action_space}A shows the results when using a clearance of $2mm$, where policies trained in all three action spaces converge in a comparable manner.
Moreover, to analyze the effect of smaller clearances, we conduct the same experiments using a $0.5mm$ clearance between peg and hole (figure~\ref{fig:results_action_space}B). However, only joint position control converges in the respective time frame. Overall, the results are different to~\\\cite{Varin2019}, where cartesian impedance control converges faster than joint position control for single-arm peg-in-hole, and~\cite{Martin-martin}, where variable cartesian impedance control performs best in contact-rich manipulation tasks. 

As peg-in-hole manipulation requires stiffness adaption, variable impedance control should theoretically perform best among the evaluated action spaces.
In our experiments this is only the case in the 2mm environment, but doesn't seem to persist when the clearance is decreased. We suspect that this is due to the increased size of the action space, which makes learning the task more challenging, but could be partially decreased by introducing isotropic gains. In the case of cartesian impedance control, we suppose that the under-performance could be attributed to the sub-optimal stiffness values. Utilizing more complex decentralized controllers comes with the downside that large effort is required for optimal parameter tuning. Hence, the results point out, that even though our framework alleviates the need to model the coupling of both manipulators manually, both decentralized controllers still require specific system knowledge. Thus, future work should investigate ways to separate the stiffness adaptation from the learning process. That way, we could have sample efficient learning while keeping the stiffness values variable.

\subsection{Robustness}
First we showcase the robustness, by evaluating disturbance recovery, and second, we demonstrate the robustness against positioning errors.\vspace{2pt}\\

\begin{figure}
	\centering
	\includegraphics[width=1\linewidth]{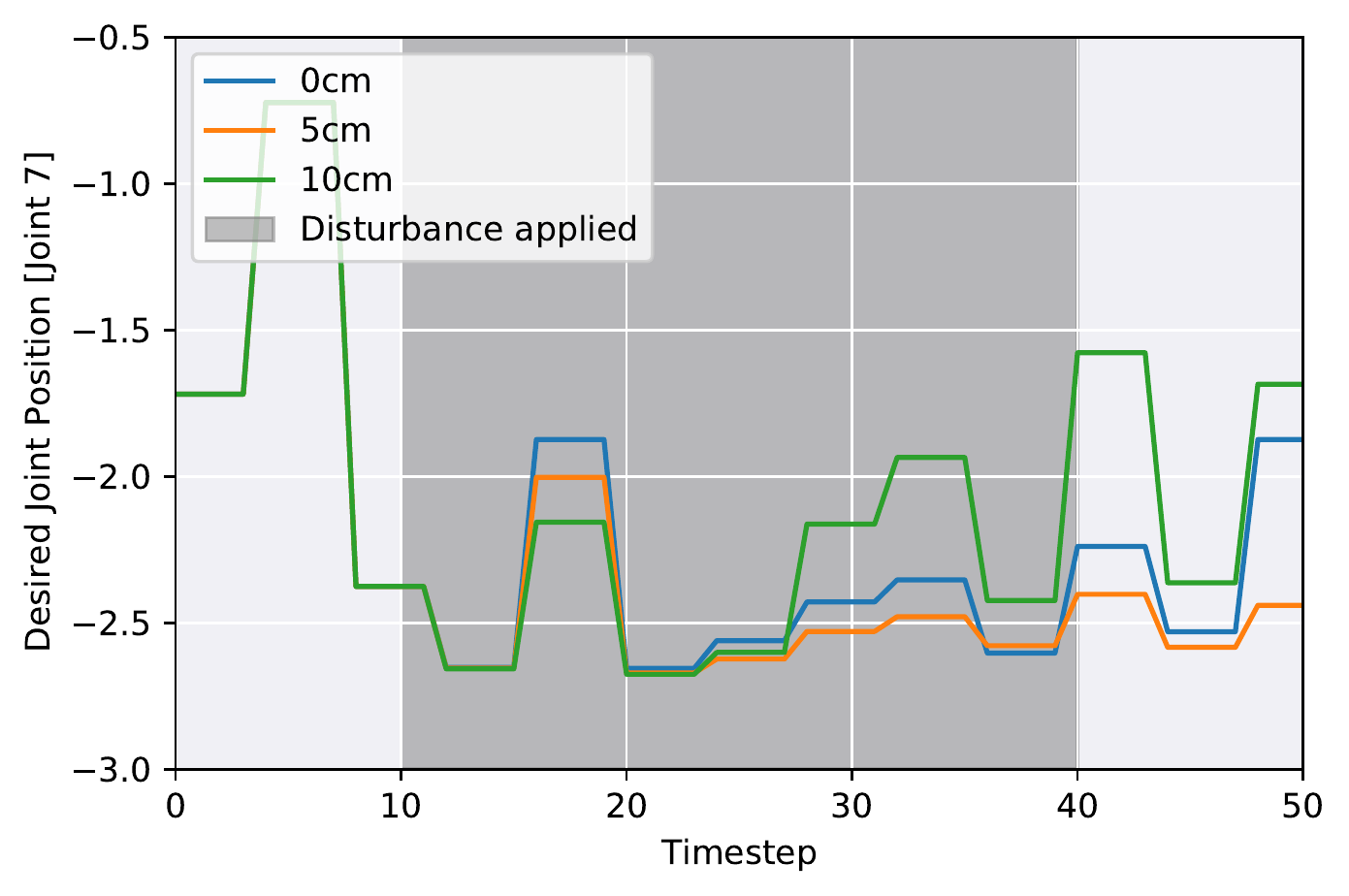}
	\caption{Desired joint positions as inferred by the policy network (joint position control) after applying a random external disturbance. }
	\label{fig:q_des}
\end{figure}

\textbf{Disturbance recovery.}
To investigate the robustness to unforeseen events, such as collision or active manipulation by humans, we evaluate the success rate after being disturbed from time step 10 till 40, resulting in an end-effector offset. Figure~\ref{fig:results_robustness}A shows the results. Each episode duration, with a maximum of 400 steps, is averaged over 60 random disturbances, 3 test cycles and all seeds. Afterwards, we compare their trajectories to a reference and calculate the disturbance as the difference between end-effector positions. Comparing all action spaces, it turns out that in our framework all variants can recover from external disturbances. Joint position control yields faster success, but episode durations increase proportionately more with higher external disturbances. Overall, the ability to recover depends mostly on the available time frame, hence increasing the maximum time steps could allow handling larger disturbance offsets. Figure~\ref{fig:q_des} visualizes trajectories of desired joint positions given by the policy network when using joint position control. 
Two time steps after the external disturbance is applied, the policy starts to adapt to the external influence, varying on the disturbance magnitude. These results show that the policy network is actively reacting to the disturbance and not purely relying on the controller.

\textbf{Position uncertainties.}
Furthermore, to evaluate the performance under position uncertainties, for instance, caused by grasping the peg in an earlier phase, we apply a random offset to the relative location of the peg to the end-effector. Figure~\ref{fig:results_robustness}B shows the success rates for three offsets. We average each result over 50 different offsets and all seeds. In general, the success rates decrease with higher offsets and smaller clearances. Cartesian impedance control turns out to be less robust compared to joint position control and variable cartesian impedance control ends up last, which is comparable to the previous robustness evaluation of disturbance recovery. Nevertheless, joint position control and cartesian impedance control are capable to handle large offsets up to $10mm$ with high success rates, which should already be sufficient for most applications and is significant considering that no randomization has been applied during training. The evaluation under positional uncertainties shows, that the policy does not simply learn the underlying kinematics since peg and hole positions are fixed during training, but infers a peg insertion behavior based on the state information.

\section{Real-World Results}
To evaluate the approach in the real-world, we transfer the policies trained in simulation without taking further measures to improve transferability such as domain randomization or domain adaption. We explicitly evaluate the performance without any additional steps targeting sim-to-real transfer, to precisely investigate if the policy is robust enough to be applied in reality and both decentralized controllers can compensate to bridge the gap between simulation and reality. 

To enable zero-shot transfer of the approach, the simulation environment has been adapted to match the real-world as close as possible. However, to ensure safety in real-world experiments, additional torque, velocity, and stiffness limitations need to be applied to guarantee observability and non-critical behavior, thereby even further increasing the sim-to-reality gap. For all real-world experiments, we use a peg and hole set with a clearance of $2mm$, since all action spaces have been successfully trained in simulation when using that clearance.

\begin{figure}
	\centering
	\includegraphics[width=1\linewidth]{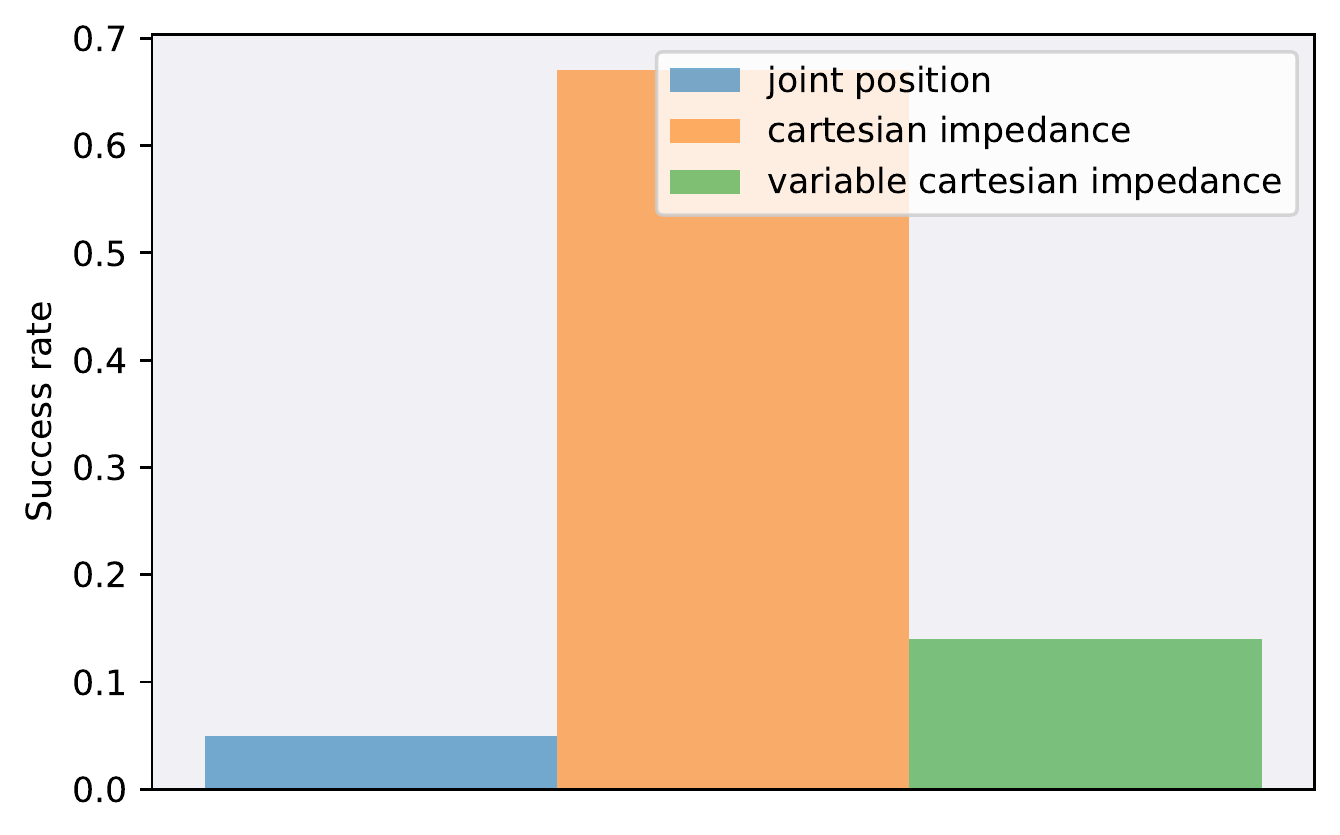}
	\caption{Success rates when transferring a policy from simulation to reality with a random deviation of $5\%$ from the initial joint positions $q_{init}$.}
	\label{fig:real_world_results}
\end{figure}

Figure~\ref{fig:real_world_results} shows the success rates for each action space when starting with a random deviation of $5\%$ for each joint from the initial joint positions $q_{init}$. Using cartesian impedance control leads by far to the highest success rate as the action space helps to compensate for the sim-to-reality gap and is robust to the applied parameter changes for safe operation. Variable impedance control confirms the results of previous robustness evaluations as the variant can not reach a high success rate in real-world evaluations. One reason for the performance decrease might be that the variable stiffness led to overfitting to the system dynamics in simulation instead of learning a generalized stiffness adaption. Joint position control, which performed best in simulation, is not able to keep up in reality at all. The action space is not robust to joint torque and joint velocity limitations, thus would require additional retraining using the applied limitations. Overall, the results show that a well-chosen action space can help to enhance robustness and transfer the approach from simulation to reality without applying further methods to target sim-to-real transfer. Moreover, the modular design helped to incorporate adaptions after training the policy, which would not have been possible in an end-to-end approach. Nevertheless, the proposed method leaves room for improvements: Among them, the impact of domain randomization and domain adaption should be explored in the future as well as fine-tuning in the real-world to adapt the policy to the additionally applied safety constraints.

\section{CONCLUSION}
We introduce a framework for dual-arm assembly with the goal to compensate for constraint and interaction modeling of traditional centralized control. The approach explores the applicability of reinforcement learning by utilizing a policy network to couple decentralized control of both robotic arms without any explicit modeling of their interaction. 
The policy is trained through model-free reinforcement learning and can be combined with various well-established single-arm controllers. As we aim to explore a framework with a minimal set of task-specific assumptions we only use sparse rewards. 
We evaluate the approach in simulation on the task of dual-arm peg-in-hole and show that joint position control provides good results up to an investigated clearances of $0.5mm$. Furthermore, we point out that in simulation the approach can recover from external disturbances and prove that the method learns a general peg insertion behavior by evaluating position uncertainties. Lastly, we zero-shot transfer the policy trained in simulation to the real-world and show that a well-chosen action space can help to overcome the sim-to-reality gap. The framework can be seen as a first step in the direction of reducing modeling efforts for dual-arm manipulation and leaves lots of room for further research including the investigation of methods to improve sim-to-reality transferability and the evaluation of further dual-arm manipulation tasks. Moreover, sample efficiency needs to be enhanced for higher precision tasks such as peg-in-hole with smaller clearances, therefore we plan to further optimize exploration, experience replay and action spaces in the future.





\section*{ACKNOWLEDGMENT}
We greatly acknowledge the funding of this work by Microsoft Germany, the Alfried Krupp von Bohlen und Halbach Foundation and project KoBo34 (project number V5ARA202) by the BMBF (grant no. 16SV7985).\\
We would like to thank Carlos Magno C. O. Valle, Luis F.C. Figueredo, Konstantin Ritt, Maximilian Ulmer and Sami Haddadin for their general support and comments on this work.


\bibliographystyle{plain}
\bibliography{bibliography}

\end{document}